\documentclass[sigconf]{acmart}

\usepackage{graphicx}
\usepackage{amsmath}
\usepackage{multirow}
\usepackage{booktabs} % for better rules in the table
\usepackage{makecell}
\usepackage{xcolor}
\usepackage{float}

\AtBeginDocument{%
  }

\copyrightyear{2025}
\acmYear{2025}
\setcopyright{cc}
\setcctype{by}
\acmConference[CSAI 2025]{2025 The 9th International Conference on Computer Science and Artificial Intelligence}{December 12--15, 2025}{Beijing, China}
\acmBooktitle{2025 The 9th International Conference on Computer Science and Artificial Intelligence (CSAI 2025), December 12--15, 2025, Beijing, China}
\acmPrice{}
\acmDOI{10.1145/3788149.3788198}
\acmISBN{979-8-4007-1962-2/2025/12}

\begin{document}

\title{Optimizing Fine-Tuning through Advanced Initialization Strategies for Low-Rank Adaptation}

\author{Yongfu Xue}
\email{xueyongfu@tongji.edu.cn}
\affiliation{%
  \institution{Tongji University}
  \city{Shanghai}
  \country{China}
}

\begin{abstract}
The rapid development of parameter-efficient fine-tuning methods has noticeably improved the efficiency of adapting large language models. Among these, LoRA has gained widespread popularity due to its strong balance of effectiveness and parameter efficiency. However, LoRA relies on initializing two low-rank matrices whose product is zero, which limits its ability to effectively activate and leverage the original model weights—creating a potential bottleneck for optimal performance. To address this limitation, we propose \mbox{\textbf{IniLoRA}}, a novel initialization strategy that initializes the low-rank matrices to closely approximate the original model weights. Experimental results indicate that IniLoRA achieves better performance than LoRA across a range of models and tasks. Additionally, we introduce two variants, IniLoRA-$\alpha$ and IniLoRA-$\beta$, both leveraging distinct initialization methods to enhance performance further. Code is available at \href{https://github.com/xueyongfu11/IniLoRA}{https://github.com/xueyongfu11/IniLoRA}
\end{abstract}



\maketitle

\section{Introduction}
Parameter-efficient fine-tuning (PEFT) methods, such as LoRA, reduce memory and computational overhead by updating only a subset of model parameters, typically through low-rank adaptations \cite{hu2021lora,valipour2022dylora,pan2024lisa,liu2024alora}.
However, research on low-rank matrix initialization remains limited. PiSSA~\cite{meng2024pissa} and MiLoRA~\cite{wang2024milora} explore different strategies for initializing low-rank matrices based on singular value decomposition (SVD). LoftQ~\cite{li2023loftq} jointly optimizes quantized weights and low-rank matrices to enhance fine-tuning performance in quantized models. Beyond these approaches, we propose several novel strategies for low-rank matrix initialization.

\begin{figure*}[ht]
    \centering
    \includegraphics[width=0.9\textwidth]{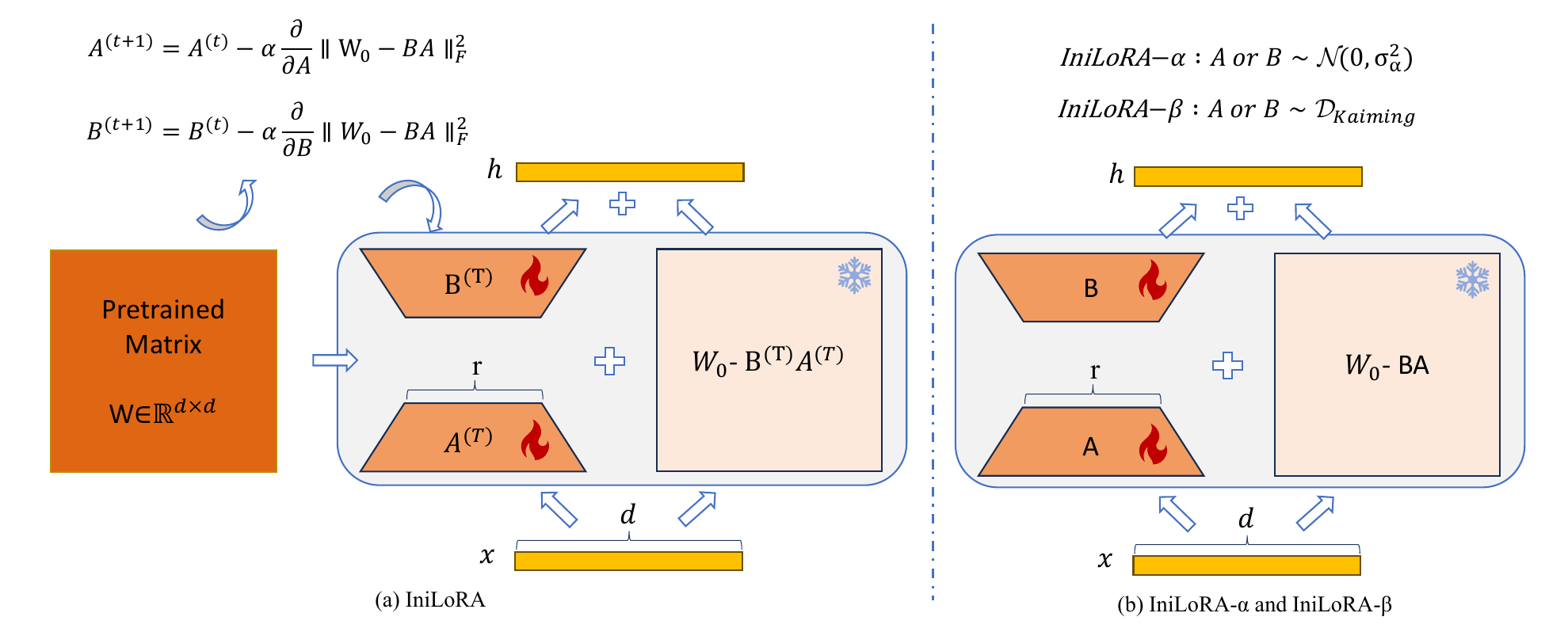}
    \caption{Our IniLoRA method and its two variants extend the LoRA framework by introducing an advanced initialization for the low-rank matrices $A$ and $B$.}
    \label{fig:architecture}
\end{figure*}

This paper introduces IniLoRA, a parameter-efficient fine-tuning (PEFT) method that approximates model weights using a product of two low-rank matrices. These matrices are optimized via gradient descent to closely match the original weights, thereby fully activating and leveraging the capacity of the pretrained model. Like LoRA, only the low-rank components are updated during fine-tuning. The difference between the original and approximated weights is captured in a residual matrix, which remains fixed during training. While prior works have explored low-rank initialization strategies \cite{li2023loftq,meng2024pissa,wang2024milora}, IniLoRA adopts a fundamentally different approach. IniLoRA leverages the differences in weight updates between full parameter fine-tuning and LoRA-based fine-tuning during the initial training phase. By aligning the initial weight adjustments with those of full parameter fine-tuning, IniLoRA facilitates convergence to a better local optimum compared to the LoRA method. IniLoRA's effectiveness is validated across benchmarks like GLUE, GSM8K, MATH, MMLU, and HumanEval.

To investigate the influence of weight initialization on performance, we propose two novel methods. First, IniLoRA-$\alpha$, which employs a broader initialization distribution, is designed to assess the role of initialization in fine-tuning. This variant consistently surpasses both standard LoRA and IniLoRA across benchmarks, highlighting the importance of initialization strategies in fully realizing the potential of parameter-efficient fine-tuning. Second, we introduce IniLoRA-$\beta$, which is based on the Kaiming distribution~\cite{he2015delving} and further explores the effect of initialization distributions on model performance.

In conclusion, the main contributions are summarized as follows:
\begin{itemize}
    \item This paper introduces IniLoRA, a novel parameter-efficient fine-tuning method. IniLoRA approximates the original model weights by utilizing the product of two low-rank matrices. 
    \item Extensive experiments were conducted across multiple models for various tasks. The results highlight IniLoRA’s robust performance.
    \item IniLoRA-$\alpha$ and IniLoRA-$\beta$ were introduced. A systematic analysis of initialization strategies with varying standard deviations and distributions was performed. The findings indicate that low-rank matrices initialized with relatively larger standard deviations and the Kaiming distribution outperform both LoRA and IniLoRA.
\end{itemize}

\section{Related work}

Parameter-efficient fine-tuning (PEFT) addresses the computational challenges of full parameter fine-tuning by adjusting only a small subset of parameters, making it more efficient for large language models. PEFT has garnered significant attention in AI research, with methods like LoRA \cite{hu2021lora} tuning low-rank matrices to approximate weight changes and merging them with pre-trained weights before inference.

\subsection{Variants of LoRA}

Zhang et al.~\cite{zhang2023adalora} used Singular Value Decomposition (SVD) to prune less important singular values for more efficient updates. Yeh et al.~\cite{yeh2023navigating} introduced LoHa, which uses Hadamard products to increase weight update ranks while maintaining the original number of trainable parameters, and LoKr, which uses the Kronecker product for the same purpose. Nikdan et al.~\cite{nikdan2024rosa} proposed RoSA, combining low-rank and sparse components to capture both primary and less prominent update directions. Other approaches include adjusting learning rates for LoRA's low-rank matrices \cite{hayou2024lora+}, mimicking full fine-tuning by decomposing weights into magnitude and direction \cite{liu2024dora}, and proving that gradients are low-rank \cite{zhao2024galore}. Ren et al.~\cite{ren2024mini} introduced MELoRA, which creates and splices small LoRAs, resembling sparse LoRA methods.

\subsection{Related Work on Low-Rank Matrix Initialization}

Several studies have enhanced LoRA's initialization. LoftQ \cite{li2023loftq} combines quantization and low-rank approximation, using SVD to better align models with pretrained weights, improving both initialization and downstream task performance. Similarly, Meng et al.~\cite{meng2024pissa} initialize low-rank matrices with principal singular values via SVD, freezing the residual matrix, and find tuning principal components more effective. MiLoRA \cite{wang2024milora} also employs SVD but updates only secondary components, reducing computational costs while preserving pretrained knowledge. Experiments show that MiLoRA outperforms LoRA and PiSSA in both performance and efficiency.

\section{Method}
\label{sec:method}

\begin{table*}[t]
    \centering
    \caption{Comparison of performance across different parameter-efficient fine-tuning methods on the GLUE benchmark using RoBERTa Base and RoBERTa Large models. The table reports average scores and parameter counts for full parameter fine-tuning (FT), BitFit \cite{zaken2021bitfit}, Adapter tuning \cite{houlsby2019parameter,lin2020exploring,pfeiffer2020adapterfusion,ruckle2020adapterdrop}, LoRA \cite{hu2021lora}, and the proposed IniLoRA method. Results show that IniLoRA achieves competitive performance compared to other methods. The symbol * indicates results reported in the LoRA paper.}
    \label{tab:glue_performance}
    \begin{tabular}{l|r|lllllll}
        \toprule
        \textbf{Model \& Method} & \textbf{\# Params} & \textbf{CoLA} & \textbf{MRPC} & \textbf{QNLI} & \textbf{RTE} & \textbf{STS-B} & \textbf{SST-2} & \textbf{Avg.} \\
        \midrule
        $\text{Ro}_\text{Base}$(FT)*       & 125M    & 63.6      & 90.2      & 92.8      & 78.7      & 91.2      & 94.8      & 85.2      \\
        $\text{Ro}_\text{Base}$(BitFit)*   & 0.1M    & 62.0      & \textbf{92.7}      & 91.8      & \textbf{81.5}      & 90.8      & 93.7      & 85.4      \\
        $\text{Ro}_\text{Base}$($\text{Adpt}^\text{D}$)*    & 0.3M    & 60.8\textsubscript{$\pm$0.4} & 88.5\textsubscript{$\pm$1.1} & 93.1\textsubscript{$\pm$0.1} & 71.5\textsubscript{$\pm$2.7} & 89.7\textsubscript{$\pm$0.3} & 94.2\textsubscript{$\pm$0.1} & 83.0      \\
        $\text{Ro}_\text{Base}$($\text{Adpt}^\text{D}$)*    & 0.9M    & 62.6\textsubscript{$\pm$0.9} & 88.4\textsubscript{$\pm$0.1} & 93.0\textsubscript{$\pm$0.2} & 75.9\textsubscript{$\pm$2.2} & 90.3\textsubscript{$\pm$0.1} & 94.7\textsubscript{$\pm$0.3} & 84.2      \\
        $\text{Ro}_\text{Base}$(LoRA)     & 0.3M    & 63.5\textsubscript{$\pm$0.8} & 88.9\textsubscript{$\pm$0.6} & 92.9\textsubscript{$\pm$0.2} & 80.3\textsubscript{$\pm$0.2} & 90.7\textsubscript{$\pm$0.1} & 94.7\textsubscript{$\pm$0.5} & 85.2      \\
        $\text{Ro}_\text{Base}$(IniLoRA)  & 0.3M    & \textbf{63.7\textsubscript{$\pm$0.9}} & 91.4\textsubscript{$\pm$0.5} & \textbf{93.8}\textsubscript{$\pm$0.1} & 79.0\textsubscript{$\pm$1.9} & \textbf{91.7\textsubscript{$\pm$0.2}} & \textbf{95.1\textsubscript{$\pm$0.2}} & \textbf{85.8}      \\
        \midrule
        $\text{Ro}_\text{Large}$(FT)*      & 356M    & 68.0      & \textbf{90.9}      & 94.7      & 86.6      & \textbf{92.4}      & 96.4      & 88.2      \\
        $\text{Ro}_\text{Large}$($\text{Adpt}^\text{P}$)*   & 3M      & 68.3\textsubscript{$\pm$1.0} & 90.2\textsubscript{$\pm$0.7} & \textbf{94.8\textsubscript{$\pm$0.2}} & 83.8\textsubscript{$\pm$2.9} & 92.1\textsubscript{$\pm$0.7} & 96.1\textsubscript{$\pm$0.3} & 87.6      \\
        $\text{Ro}_\text{Large}$($\text{Adpt}^\text{P}$)*   & 0.8M    & 67.8\textsubscript{$\pm$2.5} & 89.7\textsubscript{$\pm$1.2} & \textbf{94.8\textsubscript{$\pm$0.3}} & 80.1\textsubscript{$\pm$2.9} & 91.9\textsubscript{$\pm$0.4} & 96.6\textsubscript{$\pm$0.2} & 86.8      \\
        $\text{Ro}_\text{Large}$($\text{Adpt}^\text{H}$)*   & 6M      & 66.5\textsubscript{$\pm$4.4} & 88.7\textsubscript{$\pm$2.9} & 94.7\textsubscript{$\pm$0.2} & 83.4\textsubscript{$\pm$1.1} & 91.0\textsubscript{$\pm$1.7} & 96.2\textsubscript{$\pm$0.3} & 86.8      \\
        $\text{Ro}_\text{Large}$($\text{Adpt}^\text{H}$)*   & 0.8M    & 66.3\textsubscript{$\pm$2.0} & 87.7\textsubscript{$\pm$1.7} & 94.7\textsubscript{$\pm$0.2} & 72.9\textsubscript{$\pm$2.9} & 91.5\textsubscript{$\pm$0.5} & 96.3\textsubscript{$\pm$0.5} & 84.9      \\
        $\text{Ro}_\text{Large}$(LoRA)    & 0.8M    & 67.8\textsubscript{$\pm$1.6} & 90.1\textsubscript{$\pm$1.8} & 94.6\textsubscript{$\pm$0.2} & 87.7\textsubscript{$\pm$0.7} & 91.9\textsubscript{$\pm$0.1} & 96.5\textsubscript{$\pm$0.1} & 88.1      \\
        $\text{Ro}_\text{Large}$(IniLoRA) & 0.8M    & \textbf{69.4\textsubscript{$\pm$0.6}} & 90.0\textsubscript{$\pm$0.4} & 94.6\textsubscript{$\pm$0.3} & \textbf{88.1\textsubscript{$\pm$0.9}} & 92.1\textsubscript{$\pm$0.3} & \textbf{97.0\textsubscript{$\pm$0.1}} & \textbf{88.5}      \\
        \bottomrule
    \end{tabular}
\end{table*}

The LoRA method updates pre-trained weights \( W_0 \in \mathbb{R}^{d \times k} \) using low-rank matrices \( A \in \mathbb{R}^{r \times k} \) and \( B \in \mathbb{R}^{d \times r} \), such that \( \Delta W = BA \) and \( r \ll \min(d, k) \). Only \( A \) and \( B \) are fine-tuned via gradient descent, reducing memory use and speeding up training. IniLoRA, proposed in this paper, follows the same low-rank structure but introduces a novel initialization strategy for \( A \) and \( B \), using matrix decomposition optimized by gradient descent (as shown in Figure \ref{fig:architecture}).

To approximate the weight matrix \( W_0 \), two low-rank matrices \( A^{(0)} \in \mathbb{R}^{r \times k} \) and \( B^{(0)} \in \mathbb{R}^{d \times r} \) are initialized using a normal distribution. Rather than deriving the initialization parameters directly from \( W_0 \), a global approach is adopted for stability: the mean \( \mu_i \) and standard deviation \( \sigma_i \) of each layer’s weight matrix \( W^i \) (for \( i = 1, \dots, n \)) are computed, and their averages are used to define a unified initialization distribution for \( A^{(0)} \) and \( B^{(0)} \). This leverages the observed similarity in weight distributions across layers to streamline the initialization process.

\begin{equation}
    \mu_i = \frac{1}{| W^i |} \sum_{w \in W^i} w \; \; ; \; \; \sigma_i = \sqrt{\frac{1}{|W^i|} \sum_{w \in W^i} (w - \mu_i)^2}
\end{equation}

The average of the means and standard deviations across all layers is then computed as:
\begin{equation}
    \overline{\mu} = \frac{1}{n} \sum_{i=1}^n \mu_i \; \; ; \; \; \overline{\sigma} = \frac{1}{n} \sum_{i=1}^{n} \sigma_i
\end{equation}

The averaged values $\overline{\mu}$ and $\overline{\sigma}$ serve as global initialization parameters, setting $\mu_{\text{init}} = \overline{\mu}$ and $\sigma_{\text{init}} = \overline{\sigma}$ for all layers when initializing matrices $A$ and $B$. A low-rank approximation is then optimized by minimizing the mean squared error between $BA$ and the target weight matrix $W_0$, using gradient descent with sufficient iterations, according to the following update rules for $A$ and $B$:

\begin{equation}
    A^{(t+1)} = A^{(t)} - \alpha \frac{\partial}{\partial A} \left\| W_0 - BA \right\|_F^2  \; \; ; \; \;  B^{(t+1)} = B^{(t)} - \alpha \frac{\partial}{\partial B} \left\| W_0 - BA \right\|_F^2
\end{equation}

where $ \alpha $ is the learning rate, and $t$ represents the $t$-th iteration step. After $ T $ iterations, assuming the loss has converged, the final matrices $ A^{(T)} $ and $ B^{(T)} $ are used to initialize the low-rank matrices \(A\) and \(B\) for subsequent fine-tuning. However, even after sufficient iterations, a residual may exist between $ B^{(T)}A^{(T)} $ and $ W_0 $. To address this, we define the residual matrix $ R \in \mathbb{R}^{d \times k} $, where $ R = W_0 - B^{(T)}A^{(T)} $. During fine-tuning, only the matrices $ A $ and $ B $ are updated, while the residual matrix $ R $ remains fixed.

IniLoRA exploits differences in weight updates between full parameter fine-tuning and LoRA-based fine-tuning during the initial training phase. IniLoRA ensures that the weight adjustments during the initial training phase closely resemble those obtained through full parameter fine-tuning, as the matrix product \( BA \) closely approximates the initial weight matrix \( W_0 \). However, IniLoRA introduces an additional weight approximation step optimized using gradient descent. For any large foundational model, this weight approximation is performed only once, with the resulting matrices $ A^{(T)} $ and $ B^{(T)} $ being cached. As a result, whenever IniLoRA fine-tuning is performed, the pre-approximated weights are retrieved directly from the local cache, eliminating the need for repeated calculations before each fine-tuning session, noticeably enhancing efficiency.

When weight approximation iterations are set to zero, IniLoRA relies on randomly initialized low-rank matrices. Unlike LoRA’s zero-initialized product, this leads to a non-zero product. As shown in Table \ref{tab:zero_init} in the Appendix, IniLoRA still achieves competitive results on the GSM8K dataset under this setting with the LLaMA2-7B model, directly challenging LoRA’s initialization strategy. Motivated by this observation, we propose IniLoRA-$\alpha$, which further improves initialization by adjusting the standard deviation. Specifically, IniLoRA-$\alpha$ employs a broader initialization distribution for the low-rank matrices \( A \) and \( B \), using

\begin{equation}
A, B \sim \mathcal{N}(0, \sigma_{\alpha}^2), \quad \sigma_{\alpha} > \sigma_{\text{init}}
\end{equation}

where \(\sigma_{\alpha}\) is set to a relatively larger standard deviation (\(\sigma_{\alpha} = 0.5)\), in contrast to the standard IniLoRA initialization with \(\sigma_{\text{init}} = \overline{\sigma}\). Once the low-rank matrices \( A \) and \( B \) are initialized, fine-tuning can begin directly. This adjustment increases the variance of the initial weight distribution, allowing the low-rank matrices to explore a wider region of the parameter space.

By contrast, IniLoRA-$\beta$ employs low-rank matrices initialized using the Kaiming distribution, which better aligns with the requirements of weight initialization and gradient propagation in deep networks, stabilizes input–output variance, and demonstrates superior performance compared to both LoRA and IniLoRA.

\begin{table*}[t]
    \centering
    \caption{Performance comparison of different PEFT approaches across various tasks.}
    \label{tab:NLG_performance}
    \begin{tabular}{llcccccc}
        \toprule
        \multirow{2}{*}{\textbf{Model}} & \multirow{2}{*}{\textbf{Method}} & \multirow{2}{*}{\textbf{GSM8K}} & \multirow{2}{*}{\textbf{MATH}} & \multirow{2}{*}{\textbf{MMLU}} & \multicolumn{2}{c}{\textbf{HumanEval}} \\
        \cmidrule(lr){6-7}
                               &                           &                             &                            &                           & P@1              & P@10          \\
        \midrule
        LLaMA2-7B              & PiSSA                     & 20.4                        & 2.8                        & \textbf{44.0 }            & 14.9             & 17.3          \\
                               & MiLoRA                    & 19.1                        & 2.6                        & 43.6                      & 14.7             & 18.4          \\
                               & LoRA                      & 20.5                        & 2.4                        & 43.6                      & 14.2             & 18.3          \\
                               & IniLoRA                   & \textbf{22.5}               & \textbf{2.9 }               & 43.4                     & \textbf{15.3}    & \textbf{18.6}          \\
        \midrule
        Gemma-7B               & PiSSA                     & 65.2                        & 25.2                       & 61.0                      & 36.6             & 63.2          \\
                               & MiLoRA                    & 64.7                        & 24.4                       & \textbf{61.2}             & 36.0             & 62.4          \\
                               & LoRA                      & 64.5                        & \textbf{24.8}              & 61.1                      & 36.8             & \textbf{64.3}          \\
                               & IniLoRA                   & \textbf{66.9}               & 23.9                       & 61.1                      & \textbf{38.0}    & 64.0          \\
        \midrule
        LLaMA3-8B              & PiSSA                     & 59.7                        & \textbf{17.8}              & 60.8                       & \textbf{42.5}    & 55.1          \\
                               & MiLoRA                    & 58.0                        & 14.7                       & 60.7                       & 41.4             & 51.2          \\
                               & LoRA                      & 59.6                        & 16.3                       & \textbf{61.3}              & 40.8            & 50.0          \\
                               & IniLoRA                   & \textbf{60.1}               & 17.5                       & 60.1                       & 41.9             & \textbf{55.3}          \\
        \midrule
        LLaMA2-13B             & PiSSA                     & 37.3                        & 5.4                        & 53.1                       & \textbf{20.2}    & \textbf{25.8}          \\
                               & MiLoRA                    & 34.5                        & 5.3                        & 53.0                       & 18.7             & 24.6          \\
                               & LoRA                      & 35.9                        & \textbf{5.5}               & 53.1                       & 17.9             & 24.4          \\
                               & IniLoRA                   & \textbf{37.4}               & 5.3                        & \textbf{53.2}               & 20.1            & 25.6          \\
        \bottomrule
    \end{tabular}
\end{table*}

\section{Experiments}
\subsection{Weight approximation experiments} 

We approximate the weights of several models—RoBERTa~\cite{liu2019roberta}, LLaMA2-7B/13B~\cite{touvron2023llama}, Gemma-7B~\cite{team2024gemma}, and LLaMA3-8B~\cite{dubey2024llama}—by applying low-rank decomposition to the query and value projections in their self-attention layers. To ensure stable training, we initialize the low-rank matrices using parameters aligned with the original weight distribution, as detailed in Section~\ref{sec:method}. Following \cite{glorot2010understanding}, since the weight means are near zero but standard deviations vary, we simplify initialization by sampling from a normal distribution with zero mean.

The weight approximation experiments, applied to the query and value projections, use a learning rate of 5e-4, the Adam optimizer, and a StepLR scheduler over 20000 steps. Loss decreases rapidly early on, converging around 4000 iterations, but all experiments run the full 20000 steps to ensure adequate weight approximation across all layers.

To improve the efficiency of weight approximation and optimize computational resource usage, this experiment adopts a concurrent strategy that performs weight approximation across multiple model layers simultaneously. Table~\ref{tab:time_memory_cost} in the Appendix presents the time and memory requirements of the model evaluated in the experiment. The experiments were conducted on an NVIDIA 4090 GPU with a concurrency level of 64 and a rank value of 8. The results show that applying weight approximation to models with many layers or to self-attention layers with higher hidden dimensions substantially increases both memory consumption and computational time. In contrast, when the model employs a grouped-query attention mechanism, the number of weight parameters in the value module is considerably reduced.

\subsection{Main Results}

\subsubsection{IniLoRA: Performance on NLU}

We fine-tuned RoBERTa on six GLUE tasks—CoLA, MRPC, QNLI, RTE, STS-B, and SST-2—using consistent settings for fair comparison \cite{hu2021lora}. The model was trained for 100 epochs with AdamW (learning rate 3e-4, warm-up ratio 0.06), a batch size of 32, and max sequence length of 512, with the best result selected across the 100 epochs. We used a rank of 8 and five random seeds (0, 11, 22, 33, 44) to ensure robustness, averaging results over five runs. Fine-tuning updated low-rank matrices in the query and value layers, plus the classification head. As shown in Table \ref{tab:glue_performance}, IniLoRA delivers competitive results on RoBERTa Base while maintaining parameter efficiency.

\subsubsection{IniLoRA: Performance on NLG}

This study evaluates IniLoRA on LLaMA2-7B, LLaMA2-13B, LLaMA3-8B, and Gemma-7B across GSM8K, MATH, HumanEval, and MMLU. Models were trained for 3 epochs using the AdamW optimizer (learning rate 2e-5, warmup ratio 0.03), with a batch size of 64. Low-rank adaptation (rank 8) was applied exclusively to the query and value layers. MetaMathQA was used to train models evaluated on GSM8K and MATH, with a maximum sequence length of 512. CodeAlpaca was employed for models tested on HumanEval, also with a maximum sequence length of 512. We fine-tuned on ShareGPT for MMLU, with a maximum sequence length of 2048.

We present a comparative analysis of baseline methods, including PiSSA, MiLoRA, and LoRA, with results shown in Table \ref{tab:NLG_performance}. The findings reveal that IniLoRA shows better performance than existing PEFT methods across most tasks and models, underscoring its effectiveness in enhancing model adaptability and task-specific performance.

\subsubsection{IniLoRA-$\alpha$ and IniLoRA-$\beta$: Performance on NLG}

\begin{table}[th]
    \centering
    \small
    \caption{Performance comparison of IniLoRA-$\alpha$ and IniLoRA-$\beta$ with LoRA and IniLoRA on LLaMA2-7B and LLaMA2-13B.}
    \label{tab:var_std}
    \begin{tabular}{llccc}
        \toprule
        \textbf{Model}    &   \textbf{Method}    & \textbf{Rank 8}       & \textbf{Rank 16}  & \textbf{Rank 32}    \\
        \midrule
        \multirow{5}{*}{LLaMA2-7B}      &     LoRA      & 20.3\textsubscript{$\pm$0.2}     & 22.2\textsubscript{$\pm$0.6}     & 25.2\textsubscript{$\pm$0.4}  \\
        &     IniLoRA   & 22.4\textsubscript{$\pm$0.4}     & 25.2\textsubscript{$\pm$0.8}   & 27.5\textsubscript{$\pm$0.8}  \\
        \cmidrule(lr){2-5}
        &     IniLoRA-$\alpha$ & 28.2\textsubscript{$\pm$1.3} & 30.6\textsubscript{$\pm$1.1} & 30.9\textsubscript{$\pm$1.0} \\
        &     IniLoRA-$\beta$(K-N) & 22.8\textsubscript{$\pm$1.1} & 27.5\textsubscript{$\pm$0.9} & 29.9\textsubscript{$\pm$0.4} \\
        &     IniLoRA-$\beta$(K-U) & 23.4\textsubscript{$\pm$1.2} & 28.3\textsubscript{$\pm$1.5} &  28.1\textsubscript{$\pm$0.7} \\
        \midrule
        \multirow{5}{*}{LLaMA2-13B}      &     LoRA      & 35.9\textsubscript{$\pm$0.3}     & 37.9\textsubscript{$\pm$0.6}     & 38.2\textsubscript{$\pm$0.9} \\
        &     IniLoRA   & 37.4\textsubscript{$\pm$0.4}     & 39.2\textsubscript{$\pm$0.5}  & 40.6\textsubscript{$\pm$0.4}   \\
        \cmidrule(lr){2-5}
        &     IniLoRA-$\alpha$ & 45.6\textsubscript{$\pm$0.7} & 48.1\textsubscript{$\pm$0.9} & 46.5\textsubscript{$\pm$0.8} \\
        &     IniLoRA-$\beta$(K-N) & 38.2\textsubscript{$\pm$0.5} & 40.4\textsubscript{$\pm$0.4} & 45.0\textsubscript{$\pm$0.5} \\
        &     IniLoRA-$\beta$(K-U) & 39.7\textsubscript{$\pm$0.4} & 42.3\textsubscript{$\pm$1.0} & 43.1\textsubscript{$\pm$0.7} \\
        \bottomrule
    \end{tabular}
\end{table}

We trained on MetaMathQA and evaluated on GSM8K, with results summarized in Table \ref{tab:var_std}. IniLoRA-$\alpha$, which employs a higher initialization standard deviation (0.5), delivers notable performance improvements, consistently surpassing both LoRA and IniLoRA. IniLoRA-$\beta$ enhances low-rank adaptation by initializing both matrices with the same Kaiming distribution—either normal (K-N) or uniform (K-U). Across evaluations, both variants of IniLoRA-$\beta$ outperform LoRA and IniLoRA, achieving superior results.

\subsection{Analysis}
\subsubsection{Evaluating IniLoRA: Performance, Scalability, and Robustness}

Table \ref{tab:var_rank_performance}  in the Appendix shows that IniLoRA consistently surpasses other methods across all ranks on the GSM8K dataset with the LLaMA2-7B model, achieving the best overall performance. Figure \ref{fig:loss_merge} in the Appendix further compares the losses of IniLoRA and LoRA under different ranks, demonstrating that IniLoRA consistently maintains lower loss than LoRA.

To evaluate scalability, we trained IniLoRA on the MetaMathQA dataset with 10K, 50K, and 100K samples (one epoch) using LLaMA2-7B model, and averaged GSM8K test accuracy over multiple trials. As shown in Table \ref{tab:datasize_robustness}, both LoRA and IniLoRA improve with more data, but IniLoRA consistently achieves higher accuracy, confirming its effectiveness—especially with larger datasets and higher ranks. Further work could explore its broader scalability.

\begin{table}[h]
    \centering
    \caption{Performance comparison of LoRA and IniLoRA with different data sizes and ranks.}
    \label{tab:datasize_robustness}
    \begin{tabular}{clccc}
        \toprule
        \textbf{Data Size} & \textbf{Method}  & \textbf{Rank 8} & \textbf{Rank 16} & \textbf{Rank 32} \\ 
        \midrule
        \multirow{2}{*}{10k}  & LoRA    & 20.5   & 21.8    & 25.2    \\
                              & IniLoRA & 22.5   & 25.8    & 27.2    \\ \midrule
        \multirow{2}{*}{50k}  & LoRA    & 23.2   & 25.8    & 27.4    \\
                              & IniLoRA & 24.7   & 27.2    & 31.3    \\ \midrule
        \multirow{2}{*}{100k} & LoRA    & 26.9   & 28.6    & 30.7    \\
                              & IniLoRA & 28.0   & 32.3    & 32.5    \\ 
        \bottomrule
    \end{tabular}
\end{table}

We also tested robustness across learning rates and found IniLoRA outperforms LoRA in most cases, particularly at higher rates and ranks (Figure \ref{fig:var_lr} in the Appendix). IniLoRA demonstrates stronger adaptability and optimization across varied conditions.

\subsubsection{Impact of the degree of weight approximation on IniLoRA performance}

The fine-tuning process of IniLoRA begins with approximating the original model weights through gradient descent. A key question is: how does the degree of weight approximation affect performance? This approximation degree is quantified using the mean squared error (MSE) between \( BA \) and original weights, which serves as the loss function during gradient descent. 

\begin{figure}[t]
    \centering
    \includegraphics[width=\linewidth]{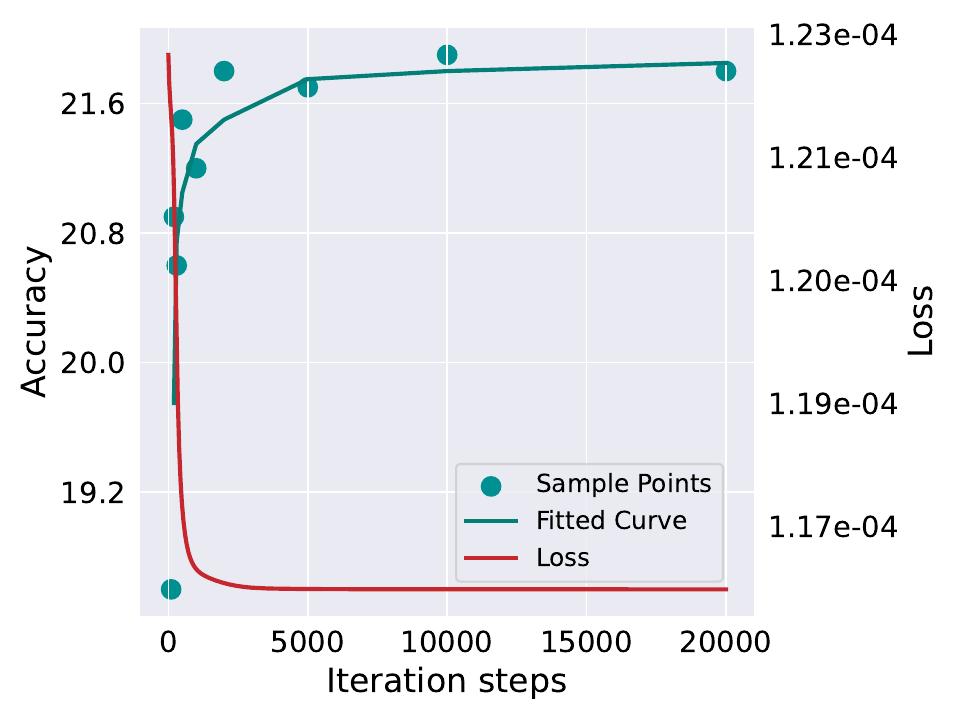}
    \caption{Correlation between performance and the degree of weight approximation in IniLoRA fine-tuning.}
    \label{fig:loss_vs_acc}
\end{figure}

We conducted experiments with the LLaMA2-7B model on MetaMathQA and evaluated downstream performance on GSM8K. Figure \ref{fig:loss_vs_acc} illustrates the relationship between weight approximation and downstream task accuracy. A reduction in loss reflects improved approximation quality. To assess IniLoRA, we examined checkpoints sampled between 100 and 20000 iterations, followed by subsequent task-specific training and validation. Each point corresponds to the performance of an individual checkpoint, while the fitted green curve denotes the moving average. The results demonstrate a pronounced decline in loss within the first 4000 iterations, yielding substantial gains in approximation and task performance. Beyond this stage, loss values plateau, suggesting that the approximation has reached a stable equilibrium with respect to performance.

\subsubsection{IniLoRA-$\alpha$: relatively larger initialization variance, better performance}
\begin{figure}[th]
    \centering
    \includegraphics[width=\linewidth]{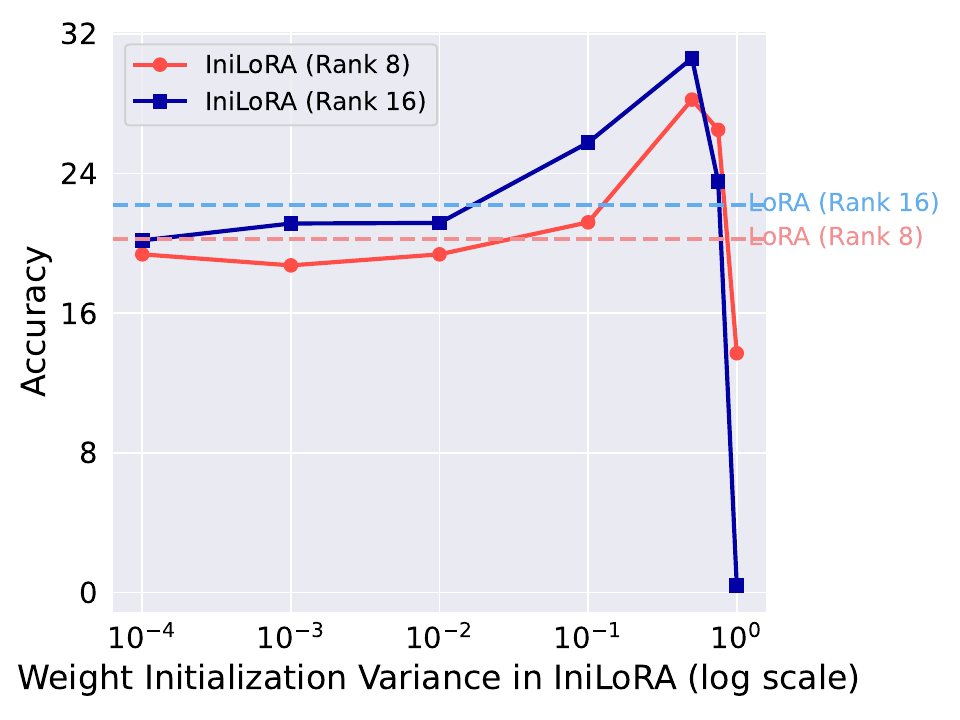}
    \caption{Impact of increasing Gaussian initialization standard deviation on model performance for LLaMA2-7B.}
    \label{fig:init_method2}
\end{figure}

Further experiments on LLaMA2-7B (trained on MetaMathQA and evaluated on GSM8K) investigate the role of initialization variance. As shown in Figure~\ref{fig:init_method2}, increasing the initialization standard deviation from 0.0001 up to 0.5 consistently enhances performance. However, when the variance exceeds 0.5, performance drops sharply, highlighting the sensitivity of model quality to initialization scale. Thus, 0.5 emerges as an optimal balance point for IniLoRA-$\alpha$.

Larger initialization weights introduce greater randomness, enabling the model to explore a broader parameter space and potentially improve performance on complex tasks. They also generate larger gradients, leading to faster updates and accelerated learning. However, proper tuning is essential—too small, and learning slows; too large, and the model may collapse.

\subsubsection{IniLoRA-$\beta$: impact of different initialization distributions}
\begin{figure}[th]
    \centering
    \includegraphics[width=\linewidth]{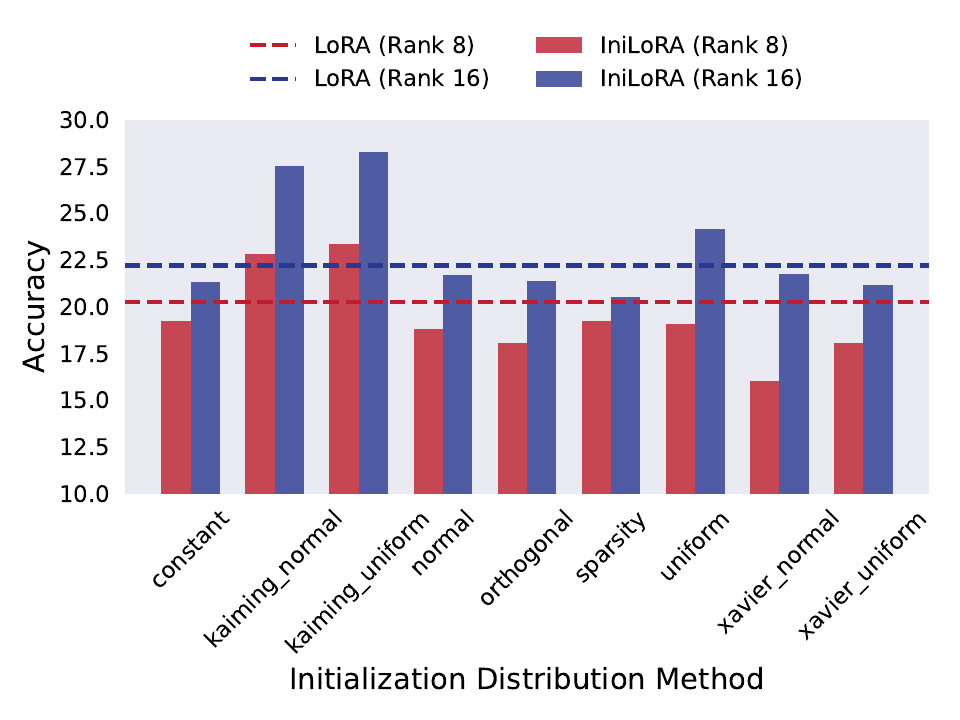}
    \caption{Performance comparison of different weight initialization methods, highlighting the consistent superior performance of Kaiming initialization.}
    \label{fig:diff_methods}
\end{figure}

We also examined the influence of different initialization distributions. Using LLaMA2-7B trained on MetaMathQA and evaluated on GSM8K, we compared multiple initialization strategies. As illustrated in Figure~\ref{fig:diff_methods}, Kaiming initialization consistently yields superior results. This indicates its effectiveness in enhancing model capacity, improving generalization, and increasing robustness against distribution shifts.

\section{Conclusion}

This paper introduces IniLoRA, a novel parameter-efficient fine-tuning method that enhances LoRA by employing a weight approximation technique. IniLoRA minimizes the mean squared error between the product of decomposed low-rank matrices and the original weights to initialize low-rank matrices. Extensive experiments on various models and tasks show that IniLoRA outperforms LoRA. In addition, IniLoRA-$\alpha$ and IniLoRA-$\beta$ are introduced. IniLoRA-$\alpha$ initializes the low-rank matrices with a normal distribution with a relatively larger standard deviation (0.5), while IniLoRA-$\beta$ employs the Kaiming initialization. Both methods outperform LoRA and IniLoRA, offering promising directions for future research in parameter-efficient fine-tuning (PEFT) techniques.

\bibliographystyle{ACM-Reference-Format}
\bibliography{custom}

\appendix

\section{Additional Results}
\label{appendix:additional_results}

\begin{table}[h]
\centering
\small
\caption{Computation time and memory usage of weight approximation in different models}
\label{tab:time_memory_cost}
\begin{tabular}{lccccc}
\hline
\textbf{Model} & \textbf{\makecell{Layers}} & \textbf{\makecell{Q/K/V\\Heads}} & \textbf{\makecell{Hidden\\ Dim.}} & \textbf{\makecell{Time \\(min)}} & \textbf{\makecell{Memory \\(GB)}} \\
\hline
RoBERTa Base   & 12 & 12, 12, 12 & 768   & 2.0   & 0.8  \\
RoBERTa Large  & 24 & 16, 16, 16 & 1024  & 4.1   & 1.3  \\
LLaMA2-7B      & 32 & 32, 32, 32 & 4096  & 25.8  & 16.2 \\
Gemma-7B       & 28 & 16, 16, 16 & 3072  & 18.2  & 12.4 \\
LLaMA3-8B      & 32 & 32, 8, 8   & 4096  & 15.7  & 10.9 \\
LLaMA2-13B     & 40 & 40, 40, 40 & 5120  & 54.6  & 24.4 \\
\hline
\end{tabular}
\end{table}

\begin{figure}[h]
    \centering
    \includegraphics[width=0.9\linewidth]{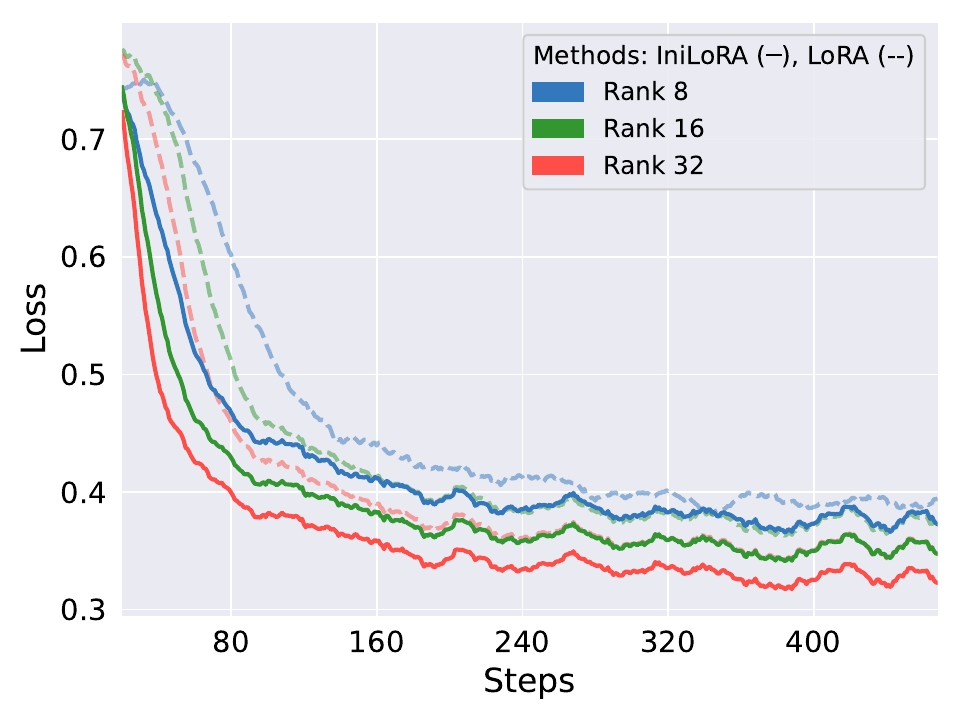}
    \caption{Comparison of training loss between IniLoRA and LoRA at various ranks, demonstrating faster convergence and lower final loss for IniLoRA.}
    \label{fig:loss_merge}
\end{figure}

\begin{figure}[h]
    \centering
    \includegraphics[width=0.9\linewidth]{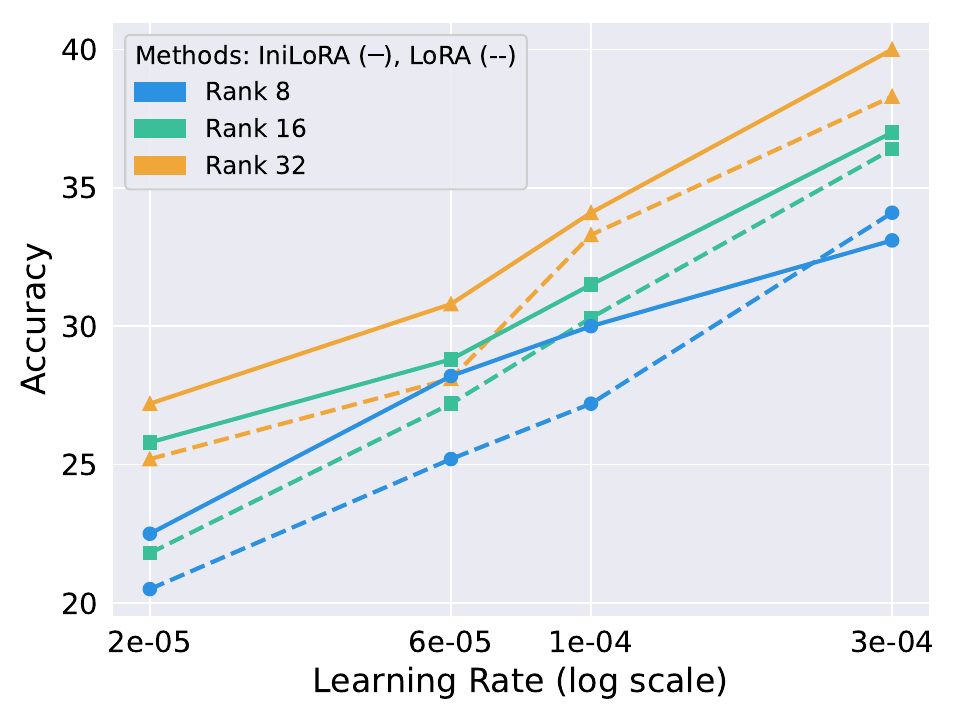}
    \caption{Impact of learning rate on IniLoRA performance.}
    \label{fig:var_lr}
\end{figure}

\begin{table}[h]
    \centering
    \caption{Performance comparison of different methods across various ranks.}
    \label{tab:var_rank_performance}
    \begin{tabular}{lccc}
        \toprule
        \textbf{Method}  & \textbf{Rank 8} & \textbf{Rank 16} & \textbf{Rank 32} \\
        \midrule
        PiSSA   & 20.4     & 24.8      & 27.2      \\
        MiLoRA  & 19.1     & 20.8      & 23.7      \\
        LoRA    & 20.5     & 21.8      & 25.2      \\
        IniLoRA & 22.5     & 25.8      & 27.2      \\
        \bottomrule
    \end{tabular}
\end{table}

\begin{table}[H]
    \centering
    \caption{Performance comparison of IniLoRA with random initialization and LoRA.}
    \label{tab:zero_init}
    \begin{tabular}{lccc}
        \toprule
        \textbf{Method}          & \textbf{Rank 8} & \textbf{Rank 16} & \textbf{Rank 32} \\ \midrule
        LoRA                     & 20.5            & 21.8             & 25.2             \\
        IniLoRA (Iter=0)         & 17.8            & 22.2             & 25.2             \\ \bottomrule
    \end{tabular}
\end{table}

\end{document}